# KP-RNN: A Deep Learning Pipeline for Human Motion Prediction and Synthesis of Performance Art

Patrick Perrine and Trevor Kirkby

California Polytechnic State University, San Luis Obispo CA 93407, USA

**Abstract.** Digitally synthesizing human motion is an inherently complex process, which can create obstacles in application areas such as virtual reality. We offer a new approach for predicting human motion, KP-RNN, a neural network which can integrate easily with existing image processing and generation pipelines. We utilize a new human motion dataset of performance art, Take The Lead, as well as the motion generation pipeline, the Everybody Dance Now system, to demonstrate the effectiveness of KP-RNN's motion predictions. We have found that our neural network can predict human dance movements effectively, which serves as a baseline result for future works using the Take The Lead dataset. Since KP-RNN can work alongside a system such as Everybody Dance Now, we argue that our approach could inspire new methods for rendering human avatar animation. This work also serves to benefit the visualization of performance art in digital platforms by utilizing accessible neural networks.

## 1 Introduction

Human motion synthesis involves digitally extracting the geometry of the human body for analysis or use in software applications. Some current research themes in human motion synthesis involve using deep neural networks for predicting human motion [1]. Such methods can have lent themselves to generating graphics for mixed reality [2]. An overarching question to our work is: how effectively can deep neural networks synthesize human motion? More specifically, a less studied and potentially more difficult motion type to learn from would be performance art. This leads us to the question: how well can deep neural networks learn from human performance art?

In this research we intended to accomplish the following:

- Implement an instance of OpenPose [3] for processing human skeletal poses from video and the Everybody Dance Now system [4] for video generation.
- Design a custom-tuned deep learning model for human motion prediction, specifically for performance art.
- Apply a new dataset to all these models.

This model could also have applications in mixed reality, computer animation, and graphics. Being able to generate new human motion and a subsequent video of predicted motion could ease the processing of human avatars. This model could also be used by dance choreographers to analyze their dance pieces. Artists sometimes are concerned with subverting or reinforcing expectations, so being able to measure the predictability of their performance could allow them greater control to work to be more predictable or less predictable in their movements.

## 2    Background

To understand human motion synthesis, one must understand the principles of computer vision. This field is concerned with utilizing digital cameras as "eyes" for computer programs to "see." If one has some intuition of computer graphics, one can conceptualize computer vision as reverse computer graphics. Whereas work in computer graphics is concerned with generating visuals based on data, computer vision is about extracting data from visuals. There has been much cross-fertilization in research for vision and graphics over decades. This has resulted in various sub-disciplines such as human motion synthesis, which has conceptual roots as early as the mid-1970's [5]. Studies in human motion saw a significant surge in the late 1990's/early 2000's [6] and have been consistently present in computer vision literature since.

Being familiar with various machine learning models, such as Recurrent Neural Networks (RNNs), Convolutional Neural Networks (CNNs), and Generative Adversarial Networks (GANs) are often required to understand the architecture of neural network models used for human motion synthesis.

Generative Adversarial Networks are systems in which two distinct neural networks are trained in competition. One of them optimizes towards distinguishing genuine and synthesized examples from a dataset, while the other optimizes towards synthesizing convincing examples to fool the first network. Generative adversarial networks have been used successfully to create or transform images, and more specifically for generating human motion [4].

## 3    Related Work

There have been various approaches to understanding human motion in digital form. A common approach is to intake video of humans performing actions, generate representations of their bodies, and then perform motion prediction [7, 8]. Deep neural networks have commonly been applied with motion prediction, sometimes in conjunction with the popular OpenPose framework [3]. Other approaches that have used graph neural networks have yielded strong motion prediction results [8]. However, by not utilizing a sequence model, such as an RNN, such methods would not lead to the capability to generate new sequences of motion, which could benefit the previously mentioned areas of virtual reality and animation.

There exists a related line of research in human motion forecasting, which focuses on predicting human movement trajectories from a distant, top-down perspective [1]. However, such methods do not account for the intricacies of human poses, rather the general direction that a given human moves within a geographic space. Motion forecasting would not lend itself as well to the visualization of full human avatars, as motion synthesis often can.

Interest in prediction methods have also lent themselves to head position prediction for virtual reality [2]. While such methods can be valuable when understanding the visual perspective of a given user in virtual reality environments, we are more concerned with entire scenes for which users could experience. Such scenes could involve the motion of various humanoid figures, rather than simply the head movements of human figures.

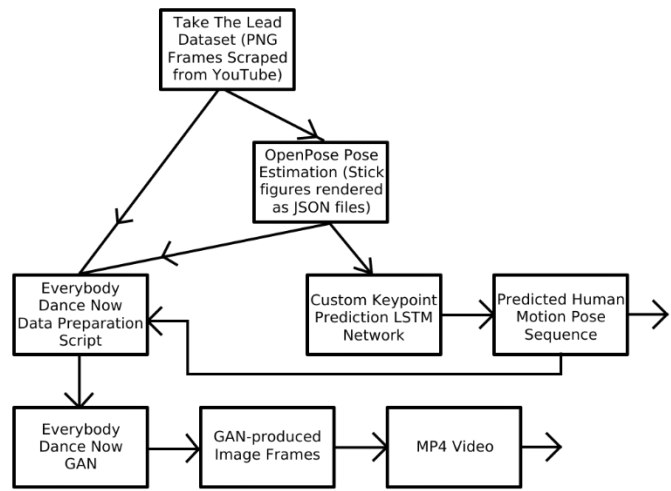

**Fig. 1.** Our video-to-video system. We can intake 2D videos of human motion and output predicted sequences of motion, as well as output synthesized video of said predicted sequences. See Figure 2 for a visual of the Everybody Dance Now GAN procedures and see Figure 3 for the Keypoint Prediction LSTM Architecture.

## 4  Data Acquisition

Our data was obtained primarily from the Take The Lead (TTL) Dataset [9]. This data was originally collated from publicly available videos of human dance posted on YouTube. Videos were downloaded and labeled by hand to reflect several different genres of dance. We acknowledge that there may exist some heterogeneity within the image data, due to inconsistent video capture procedures. This inconsistency could affect results drawn from models trained on such data. Sample data from the Everybody Dance Now system was also used as a reference point. This includes footage of collaborators in the Everybody Dance Now project performing basic human motions optimized for testing the Everybody Dance Now system. The sample data used in the previous experiments is linked publicly in the associated repository here: https://github.com/carolineec/EverybodyDanceNow. There were no human subjects involved within the course of this work.

As with virtually all machine learning systems, the products of our system may inherit biases present in the data the system is trained upon. This issue has continued to be addressed within the academic communities of machine learning [10]. The potential of bias is difficult to eliminate from machine learning, and it is important to consider the potential biases introduced through datasets. A significant mitigating factor of bias in this work is that all video of humans is preprocessed down to simple "stick figures", which only convey information about a subject's physical pose. Since our system is only aware of humans as stick figures, we believe it is difficult for our system to inherit biases from information such as race, gender, et cetera. The datasets we used were also acquired from two existing works that were vetted prior to publication. We suggest the investigation of ethical frameworks when creating new datasets to be released publicly [10].

## 5    System Design

Our system (see Figure 1) learns from the contents of YouTube videos of dancing, obtained from Take The Lead. These are stored as sequences of PNG images for each video frame. From there, the OpenPose system is used to estimate the human poses in each frame, creating a skeletal representation of the human in a given frame. The skeletal representation is translated into a JSON format that gives two-dimensional coordinates for various keypoints (head, chest, shoulders, et cetera). At this point, the data may be used in two separate models: the Everybody Dance Now system and our Long Short-Term Memory (LSTM) neural network.

To make the Everybody Dance Now (EDN) system work for videos scraped from YouTube, which often contain multiple people, the OpenPose JSON files are altered to only keep track of the poses for a single person in each video. Then a data preparation script provided in the EDN repository is used to convert the JSON files to images of stick figures, which are provided as inputs to EDN's conditional GAN. EDN trains from footage of a specific person performing movements and can then create footage of that same person performing new movements. The footage is provided as sequential image frames, which are resized to a correct aspect ratio and then combined into a single video.

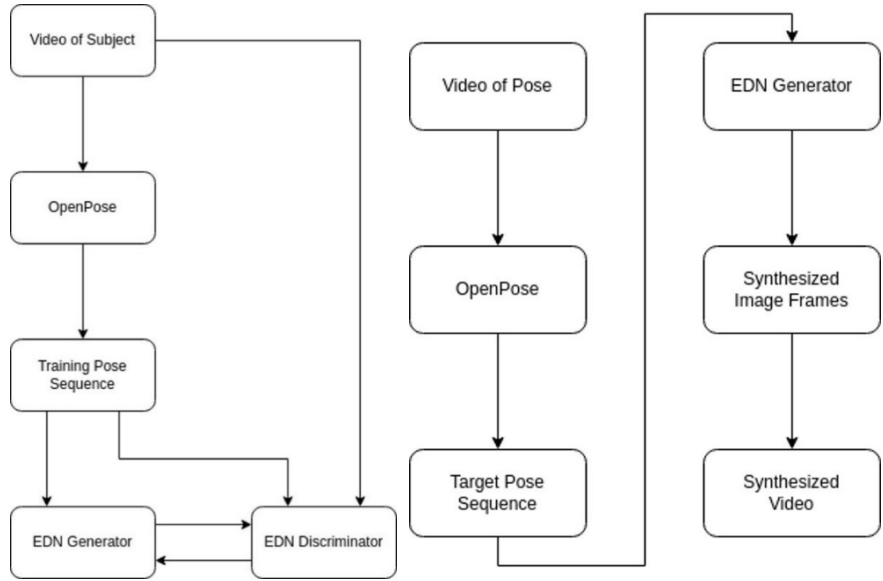

**Fig. 2a and 2b.** a) The Training Procedure for Everybody Dance Now. b) The Generation Procedure for Everybody Dance Now.

The custom keypoint predicting LSTM network, (KP-RNN, seen in Figure 3) accepts a sequence of poses as input. This input is represented as 50 numbers, which are the *(x,y)* coordinates of the 25 keypoints created by OpenPose to model the position of a single person. It has several recurrent layers and two densely connected layers at the end. We use conventional stochastic gradient descent to optimize mean squared error. The output of KP-RNN is the predicted pose in the next frame of the video for the person, again represented as 50 numbers. It should be noted that the predictions of this network can be used to generate a sequence of poses, which can then be provided as an input to Everybody Dance Now to generate a new video. There was some fine-tuning of the architectural design of KP-RNN, but not so much as to overfit our data.

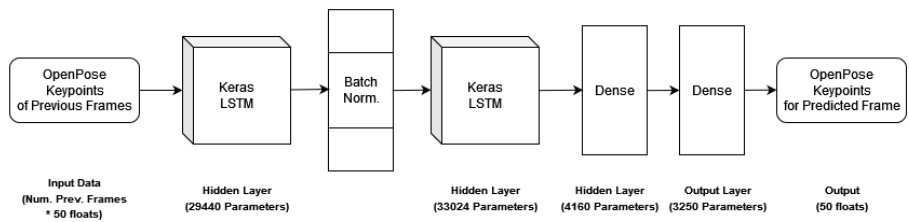

**Fig. 3.** Our Keypoint Recurrent Neural Network Architecture (KP-RNN).

### 5.1 Implementation

This system was set up on an instance of Google Colab Pro running with a Tesla P100 GPU as a development environment. The system is implemented using primarily Python and Bash, and both Everybody Dance Now and the custom architecture KP-RNN are implemented using Tensorflow [11] and Keras [12]. Results are visualized with Tensorboard and Python matplotlib, and pre-processed into video using Python Image Library and the FFMPEG command-line interface. Our parameter settings are described in Table 1.

| Parameter | Setting |
|---|---|
| Dimensions of Input Vector | 25 x 2 |
| Dimensions of Output Vector | 25 x 2 |
| Number of LSTM Layers | 2 |
| Size of LSTM Layers | 64 |
| Number of Dense Hidden Layers | 1 |
| Size of Dense Hidden Layers | 64 |
| Activation | Sigmoid |
| Dropout Probability | 0.3 |
| Loss Function | MSE |
| Optimizer | SGD |
| Learning Rate | $3 \times 10^{-3} - 1 \times 10^{-3}$ |
| Momentum | 0.2 |
| Maximum Epochs | 1200 |

**Table. 1.** KP-RNN Hyperparameter Description.

## 6 Testing and Validation

Our experiment requires measuring several different metrics to gauge the success of different components of the system.

First, we evaluate the efficacy of the KP-RNN architecture using root mean squared error. Root mean squared error is measured as: the distance between the predicted set of keypoint coordinates and the actual set of keypoint coordinates; a low root mean squared error means that the model is predicting a sequence of movements that are close to the ground truth. The other potential measurement we consider is prediction accuracy; however, for evaluating KP-RNN, root mean squared error is deemed a more useful metric than prediction accuracy. Given that motion prediction is a form of regression task, KP-RNN is being applied to a regression task and not a classification task. Prediction accuracy scores each case in an all-or-nothing manner, but this removes information compared to root mean squared error and is therefore less relevant to the problem space. For example, when using prediction accuracy, a prediction that is significantly different from the ground truth is treated the same as a prediction that is completely correct except it is offset by one pixel from the ground truth. For this reason, a low root mean squared error is our best indication of success for predicting movements.

$$\text{RMSE} = \sqrt{(y - \hat{y})^2}$$

**Eq. 1.** Formula for Root Mean Squared Error

We evaluate the success of the EDN component of our system based on scoring the difference between the generated video and the actual ground truth video. This requires scoring the visual similarity between two images that are not identical, which is not a trivial task. There are several metrics that are used for this purpose. The first such metric is the feature matching loss (figures 4a and 5a). Feature matching is calculated by running both the generated image and the actual image for a given pose through the image discriminator neural network, comparing the similarities between the vector outputs of intermediate layers in the discriminator network, and optimizing to improve this similarity [13]. The second metric is perceptual reconstruction loss (figures 4b and 5b), which takes a similar approach, except instead of using the intermediate layers of the GAN discriminator, it observes the intermediate layers of a completely different image recognition neural network architecture, specifically VGG19 [5]. Note that, for this metric, EDN records the differences between intermediate layers rather than the similarities, so while a higher value indicates better accuracy for feature matching, the opposite is true for perceptual reconstruction loss. In conclusion, a high feature matching value and a low perceptual reconstruction loss implies that the sequence of images synthesized by the system are visually similar to the ground truth images, which indicates that the system is working as intended.

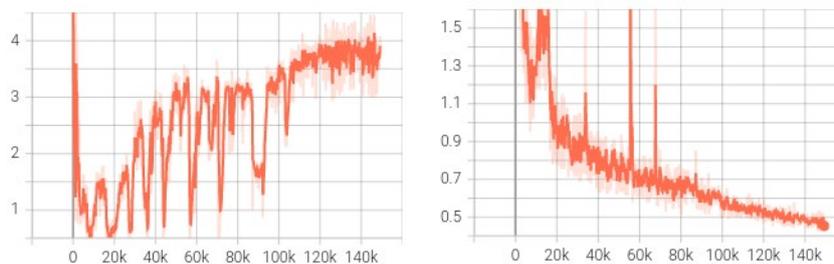

**Fig. 4a and 4b.** a) EDN Feature Matching with its own data. b) EDN Perceptual Reconstruction Loss with its own data.

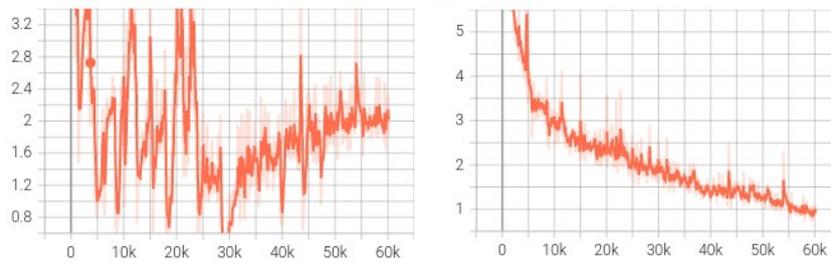

**Fig. 5a and 5b.** a) EDN Feature Matching with TTL data. b) EDN Perceptual Reconstruction Loss with TTL Data.

Recording the training loss values for the actual generator and discriminator neural networks of EDN is comparatively not useful for the purposes of evaluation. By construction, they exist in a zero-sum game, and in a successful GAN they will remain approximately matched. Because the loss metrics for the GAN, by design, contain virtually no meaningful insight into whether the system is working or not, they are omitted as evaluation metrics in this work.

Finally, EDN also produces video outputs, which may be evaluated qualitatively. See figures 6b and 7b for sample frames from the generated video, based on the inputs in figures 6a and 7a. Figure 6a is from the Everybody Dance Now demonstration dataset, and uses a more complex skeleton figure with individual keypoints for hands and face, whereas figure 7a is from the Take The Lead dataset, and uses a simplified skeleton with fewer keypoints.

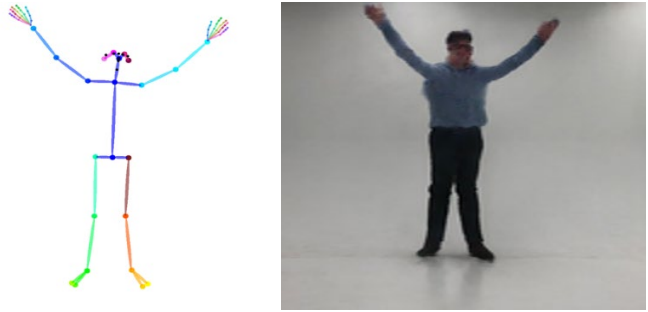

**Fig. 6a and 6b.** a) Skeleton input image from EDN's data. b) Synthesized reconstructed image from EDN's GAN.

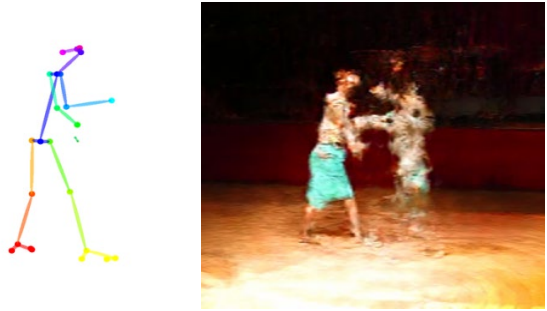

**Fig. 7a and 7b.** a) Skeleton input image from TTL's data. b) Synthesized reconstructed image from EDN's GAN.

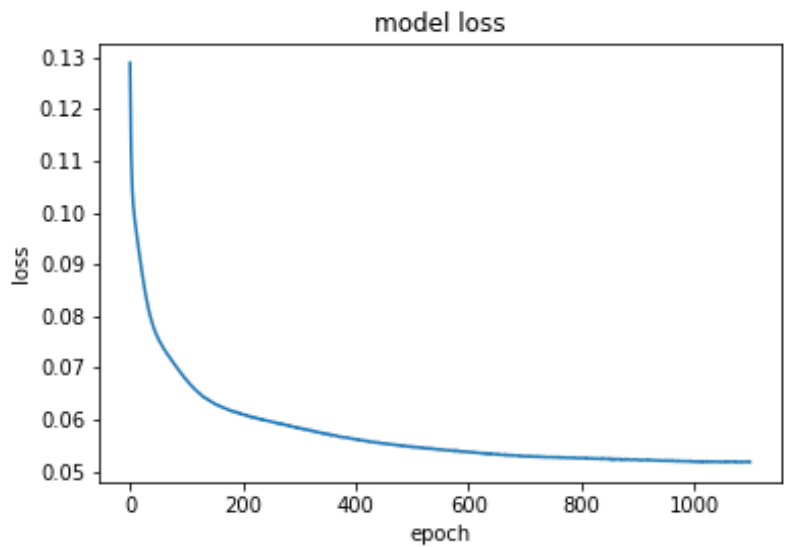

**Fig. 8.** KP-RNN training error over time.

### 6.1 Project Goal Evaluation

Here is how we accomplished our original goals, along with some technical details:

- Implement an instance of OpenPose [3] for processing human skeletal poses from video and the Everybody Dance Now system [4] for video generation.

This was mostly accomplished. However, for EDN, we observed an issue where the model does not recognize more than one person, which posed an issue when using the TTL dataset. We built on the previous work by implementing a method to pre-process new videos of multiple people and focus on a single person throughout the video se-quence. This complies with the limitation of EDN being only able to process one person's sequence.

- Design a custom deep learning model for human motion prediction, specifically tailored for performance art.

We now have our own LSTM-based architecture, KP-RNN, for predicting human motion sequences, which we developed in Google Colab Pro. Our final root mean squared error was 0.2328, which indicates that our system is functional. Given that we are using a new dataset of different structure to commonly studied motion datasets, it would be difficult to compare our error score with other prediction methods. This work can serve as a basis for future work to develop new methods for this dataset to compete with this score. Our code is publicly available on this GitHub repository: https://github.com/patrickrperrine/comp-choreo.

- Apply a new dataset to all these models.

We were interested in creating a new dataset of performance art to be used for prediction along with our existing TTL dataset. We found this to be difficult and too time-constrained for the scope of this project.

## 7    Future Work

We found that creating a new, usable dataset can be rather difficult. Acquiring new performance art data as originally proposed could be a fruitful endeavor. Also, having more computational power to try to build larger, custom architectures could lead to novel results in human motion prediction. To move past 2D motion, there could be an exploration of the Human 3.6M dataset [14], which is popular in 3D human motion prediction models [7]. Exploring newer, related topics such as human trajectory prediction [1] using transfer learning could be of interest.

## 8    Conclusion

We offer a novel approach to human motion prediction with our lightweight neural network, and its ability to integrate nicely with existing image processing pipelines. We have effectively combined EDN and TTL along with our own deep neural network to produce a new system for dance motion prediction, image-to-image translation, and video generation. Our results indicate that our LSTM-based prediction network functions effectively on a new video dataset of human performance art. Our overall processing system could inspire innovations in spaces, such as virtual reality, that are concerned with visualizing complex forms of human motion.

# 9 Acknowledgements

This work was partially supported by a Cal Poly Graduate Assistant Fellowship to Patrick Perrine. We thank Professors Franz Kurfess and Jonathan Ventura for their support of this work.